\documentclass[runningheads]{llncs}
\usepackage[T1]{fontenc}

\usepackage{amsmath,graphicx}
\usepackage{xcolor}
\usepackage{enumitem}
\setlist{nosep, leftmargin=14pt}

\usepackage{mwe} 

\usepackage{multirow}
\usepackage{booktabs} 
\usepackage{comment}
\usepackage{url}
\usepackage{array}
\usepackage{float}
\usepackage{tabularx}
\usepackage{hyperref}

\newcommand\blfootnote[1]{%
  \begingroup
  \renewcommand\thefootnote{}\footnote{#1}%
  \addtocounter{footnote}{-1}%
  \endgroup
}

\begin{document}

\title{StrokeSeg2: Stroke Lesion Segmentation in Clinical Research Workflows}

\titlerunning{StrokeSeg2}
%
\author{Youwan Mah\'e\inst{1,2*}\orcidID{0009-0006-9289-7178} \and\\
Axel Plessis\inst{1*}\orcidID{0009-0002-8154-3092} \and \\
St\'ephanie Leplaideur\inst{1,3,4}\orcidID{0000-0003-3859-4644} \and\\
Elise Bannier\inst{1, 5}\orcidID{0000-0002-8942-7486} \and\\
Florent Leray\inst{1\dagger}\orcidID{0009-0005-1062-2764} \and\\
Francesca Galassi\inst{1\dagger}\orcidID{0000-0002-8788-2856}}
\institute{Univ Rennes, Inria, CNRS, Inserm, IRISA UMR 6074, Empenn, Rennes, France\and
Siemens Healthineers, Courbevoie, France \and
CHU Rennes, Physical Medicine and Rehabilitation Department, Rennes, France \and
Centre de Kerpape, Ploemeur, France \and
CHU Rennes, Radiology Department, Rennes, France\\
\email{youwan.mahe@inria.fr}}

\authorrunning{Y. Mah\'e, A. Plessis et al.}

\maketitle

\begin{abstract}
Deep learning frameworks like nnU-Net achieve state-of-the-art brain lesion segmentation performance but remain difficult to deploy in clinical research environments due to, among other reasons, software dependencies and computational requirements. We introduce StrokeSeg2, a lightweight, modular, cross-platform C++/Qt  framework designed to adapt resource-intensive 3D stroke segmentation pipelines into portable and reproducible applications. To improve compatibility with standard clinical workstations, we investigate the combined effect of architectural compression through knowledge distillation and inference optimisation using ONNX Runtime with \texttt{Float16} quantisation. Across heterogeneous hardware configurations (CPU, integrated GPU, and dedicated GPU) architectural distillation emerged as the primary contributor to efficiency gains, contributing to over 90\% reduction in energy consumption and an average 84\% reduction in inference time. Specifically, we identify a 0.84M-parameter student model as the most favourable trade-off, reducing the original 102.3M-parameter teacher architecture to a 2.1~MB disk footprint while preserving robust lesion localisation and competitive segmentation performance. This small footprint supports the development of a self-contained installer for clinical workstation targets. Finally, StrokeSeg2 packages these optimisations into standalone installers for Windows, macOS, and Linux. By providing both graphical and command-line interfaces without Docker or external environment dependencies, StrokeSeg2 facilitates deployment of high-performance segmentation workflows for routine clinical research pipelines.\blfootnote{* indicates joint first authors and $\dagger$ indicates joint last authors.}

\keywords{stroke lesion segmentation \and nnU-Net \and ONNX Runtime \and clinical deployment \and distillation \and quantisation \and software engineering}
\end{abstract}

\section{Introduction}

\label{sec:intro}

Stroke is a leading cause of long-term disability worldwide~\cite{feigin2021global}. In the sub-acute and chronic phases, structural magnetic resonance imaging (MRI) is routinely used to assess lesion extent, monitor recovery, and guide rehabilitation~\cite{bernhardt2017agreed}. Accurate lesion segmentation is essential for quantifying brain damage and studying links between imaging biomarkers and functional outcomes.
Recent deep learning approaches, notably nnU-Net~\cite{isensee2021nnunet}, have achieved state-of-the-art performance for post-stroke lesion segmentation on datasets such as ATLAS~v2.0~\cite{liew2018atlas}. Di~Matteo~et~al.~\cite{dimatteo2025stroke} further trained stroke-specific nnU-Net~v2 models on T1-weighted (T1w) and T1w+FLAIR MRI. 

Despite these advances, the practical adoption of such models in clinical research environments remains limited. Existing computational tools generally fall into two categories: research-oriented preprocessing libraries (e.g., TorchIO~\cite{torchio2021}, pymia~\cite{jungo2020pymia}) and deployment frameworks designed for integration with clinical infrastructures (e.g., MONAI~Deploy~\cite{monai_deploy}, PACS-AI~\cite{theriault2024pacsai}). While these solutions support reproducibility and system integration, they are not primarily designed as lightweight standalone applications that can be easily deployed on standard clinical research workstations. In practice, many state-of-the-art segmentation pipelines still require complex software environments and specialised hardware, limiting use outside specialised computational settings.

Preliminary work introduced a first lightweight and portable version of a stroke lesion segmentation framework compatible with nnU-Net~\cite{previouswork2025}. In the present work, we extend this framework by investigating how 3D medical image segmentation pipelines can be compressed and adapted to constrained deployment environments while preserving clinically relevant segmentation behaviour. In particular, we investigate the impact of model distillation, \texttt{Float16} (FP16) quantisation, and heterogeneous inference backends on segmentation quality, runtime, and energy consumption across CPU-only, integrated GPU (iGPU), and dedicated GPU (dGPU) configurations.

\section{Methods}
\label{sec:methods}

\subsection{Architecture overview}
We transformed the stroke-specific nnU-Net–based segmentation pipeline of Di Matteo et al.~\cite{dimatteo2025stroke} into a lightweight, portable, and reproducible framework for clinical research workflow. 

The framework was designed to support two user groups: (i) clinical researchers through a simple and self-contained installation workflow, and (ii) computer vision researchers through a transparent and modular architecture that facilitates further development. Bridging these domains - computer vision research, software engineering, and clinical research workflows - introduces specific technical and usability constraints. To address them, development was structured along three axes:~(1)~modular architecture separating preprocessing, inference, and postprocessing; (2) performance optimisation through runtime and model-level improvements; and (3) packaging for reproducible cross-platform deployment. Deployment targeted Windows, macOS and Linux-based systems, reflecting typical usage scenarios: Windows/macOS systems for clinical research applications, and Linux systems for computer vision research and high-performance computing (HPC) environments.

The refactored nnU-Net–based pipeline~\cite{isensee2021nnunet,dimatteo2025stroke} comprises three stages: BIDS-compliant preprocessing including spatial normalisation and brain extraction using the Anima toolbox~\cite{anima_empenn}; inference via ONNX Runtime to eliminate heavy training dependencies while retaining patch-based Gaussian merging; and postprocessing, where logits ($z$) are converted via softmax ($\sigma$) into probability maps and binary masks (default threshold $0.5$), with optional spatial transforms to recover subject-space results. We take advantage of the execution abstraction provided by ONNX Runtime or Windows ML to automatically detect and select the optimal device among dedicated GPUs (NVIDIA/AMD), integrated GPUs, and the CPU for inference.

\subsection{Deployment-oriented model compression}
Deploying 3D nnU-Net segmentation pipelines in constrained clinical research environments requires reducing both model complexity and inference cost. To address this challenge, we investigated the combined impact of architectural compression through knowledge distillation~\cite{Hinton2015} and inference optimisation through reduced-precision computation.

Model distillation was implemented by extending the standard nnU-Net training framework used in~\cite{dimatteo2025stroke}. Training was performed on 655 subjects from the ATLAS~v2.0 dataset, while evaluation used 300 independent subjects from the ATLAS~v2.1 dataset. To study the trade-off between segmentation performance and computational efficiency, we designed nine student architectures by progressively reducing the number of features per stage in the original nnU-Net model. 
During training, the pre-trained teacher model remained frozen while the student network learned from both ground-truth annotations and teacher predictions for 1000 epochs (batch size of 6). The optimisation objective combined equally weighted supervised segmentation and distillation losses, the latter based on the Kullback-Leibler divergence between softened teacher and student logits~\cite{Hinton2015} ($\alpha = 0.5$). A temperature scaling factor of $T=4.0$ was applied prior to the softmax activation~:

\begin{equation}
    \label{eq:loss}
    \mathcal{L}_{\text{total}} = (1 - \alpha) \mathcal{L}_{\text{hard}} + \alpha T^2 \cdot \mathcal{D}_{\text{KL}}\left( \sigma\left(\frac{z_t}{T}\right) \Bigg\| \sigma\left(\frac{z_s}{T}\right) \right)
\end{equation}

Training was performed on NVIDIA H100 GPUs. In addition to architectural compression through distillation, ONNX \texttt{Float16} quantisation was applied during inference to further reduce model size and memory requirements. 

We evaluated the execution performance of distilled and quantised models across three hardware configurations representative of typical clinical research environments: (1) CPU-only execution (Intel\textsuperscript{\textregistered} Core \textsuperscript{TM} Ultra 7 265HX), (2) integrated GPU execution (Intel\textsuperscript{\textregistered} Graphics), and (3) dedicated GPU execution (NVIDIA RTX Pro 1000 Blackwell). 
Three hundred chronic stroke subjects from the ATLAS v2.1~\cite{liew2018atlas} dataset were processed using the monomodal T1w pipeline with FP32 and FP16 ONNX models. Segmentation quality was assessed using global Dice score, lesion-wise Dice score and average surface distance (ASD), using a fixed probability threshold of 0.5 to generate binary lesion masks. Lesion detection performance was measured using lesion-wise F1 score, considering lesions with at least 10\% overlap relative to ground-truth lesion volume as true positives. Metrics were computed both globally and after stratification by lesion size (small, medium, and large), using the first and third quartiles of lesion volume as thresholds.

\subsection{Packaging and deployment}

\begin{figure}[]
  \centering
  \includegraphics[width=\linewidth]{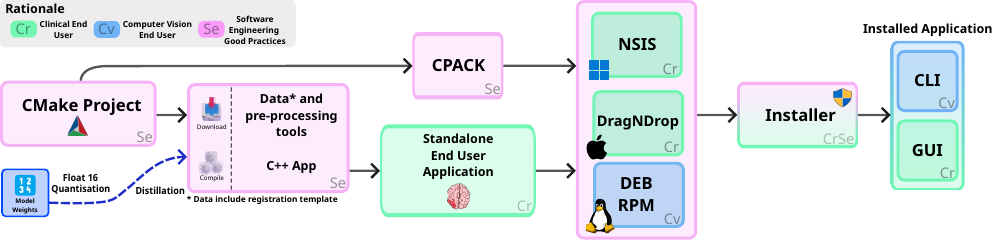}
  \caption{StrokeSeg2 software engineering and distribution workflow. Model weights undergo distillation and \texttt{Float16} quantisation before being integrated into a modular C++ application. The framework utilises CPack to generate OS-specific installers (NSIS, DragNDrop, DEB/RPM)  for clinical and research deployment.}
  \label{fig:framework}
\end{figure}

For all platforms, StrokeSeg2 is packaged as a self-contained C++/Qt application using CMake/CPack to generate platform-specific installers. For Windows, CPack uses NSIS to generate an NSIS installer. On macOS, we deliver dmg thanks to the DragAndDrop CPack generator. Finally, CPack generates DEB or RPM to target Debian and Red Hat-based installations, respectively~(Fig.\ref{fig:framework}).

Self-containment is essential for clinical end users, who typically have limited interaction with command-line tools. The installers bundle all required dependencies. This fully contained build ensures reproducibility, as all package versions are fixed at compile time. The resulting package supports OS-compliant installation and uninstallation procedures, providing clean integration and removal.

\subsection{Software architecture and extensibility}
The framework relies on ONNX Runtime as a unified inference backend, allowing trained 3D nnU-Net models to be integrated after ONNX conversion without modifying the application backend. This design facilitates adaptation to additional MRI modalities and lesion segmentation tasks while preserving the same deployment infrastructure.

The codebase is organised into two repositories: (i) the application, (ii) the build and packaging scripts using CMake/CPack. The latter provides a simple, dependency-free build process that automatically bundles all required components. This modular structure facilitates integration into broader clinical or research pipelines, ensuring transparency, portability, and maintainability while providing a lightweight alternative to Docker-based deployment. Additional features include probability map export, configurable lesion thresholds, and optional MNI-space outputs. Application data, including logs, configuration files, and model storage, follows standard operating-system conventions.

\section{Results}
\subsection{Performance and size}
\label{sec:KD}

\paragraph{}
We developed a series of student networks by progressively reducing the number of feature channels at each stage, ranging from the original 102.3M-parameter teacher model down to a compact 0.05M-parameter variant~(Table~\ref{tab:model_equivalence}). To optimise computational throughput and GPU memory alignment, feature counts were kept divisible by 16 where feasible.

\begin{table}[htb]
\centering
\caption{Parameters and storage footprint of teacher (T) and distilled student networks~(Sx). The table reports parameter counts, checkpoint size across formats (PyTorch FP32, ONNX FP32, ONNX FP16), and feature dimensions per stage. Reduction ratios compare each original PyTorch FP32 checkpoint with its ONNX FP16 counterpart.}
\label{tab:model_equivalence}
\footnotesize
\begin{tabularx}{\textwidth}{l *{5}{>{\centering\arraybackslash}X} *{6}{r}}\toprule
\textbf{Model} & \textbf{Params (M)} & \textbf{PyTorch FP32 (MB)} & \textbf{ONNX FP32 (MB)} & \textbf{ONNX FP16 (MB)} & \textbf{Red. Ratio} & \multicolumn{6}{c}{\footnotesize\textbf{Features per stage}} \\
\midrule
\textbf{T} & 102.3 & 820 & 411 & 206 & 75\% & 32 & 64 & 128 & 256 & 320 & 320 \\
\midrule
\textbf{S1}      & 52.9  & 424 & 213 & 107 & 75\% & 28 & 56 & 112 & 224 & 320 & 320 \\
\textbf{S2}      & 35.2  & 283 & 142 & 71  & 75\% & 24 & 48 & 96  & 192 & 256 & 256 \\
\textbf{S3}      & 16.3  & 132 & 66  & 33  & 75\% & 20 & 40 & 80  & 160 & 200 & 200 \\
\textbf{S4}      & 10.4  & 84  & 43  & 21  & 75\% & 16 & 32 & 64  & 128 & 160 & 160 \\
\textbf{S5}      & 4.7   & 38  & 19  & 9.9 & 74\% & 12 & 24 & 48  & 80  & 96  & 128 \\
\textbf{S6}      & 2.6   & 21  & 11  & 5.7 & 73\% & 8  & 16 & 32  & 64  & 80  & 80  \\
\textbf{S7}      & 0.84  & 7.2 & 3.8 & 2.1 & 71\% & 4  & 8  & 16  & 32  & 48  & 48  \\
\textbf{S8}      & 0.21  & 2.2 & 1.2 & 0.8 & 64\% & 2  & 4  & 8   & 16  & 24  & 24  \\
\textbf{S9}      & 0.05  & 0.9 & 0.6 & 0.5 & 44\% & 1  & 2  & 4   & 8   & 12  & 12  \\
\bottomrule
\end{tabularx}
\end{table}

The transition from the native PyTorch implementation to ONNX runtime, followed by \texttt{Float16} quantisation, yielded significant reductions in model size, with ONNX FP16 checkpoints reducing storage requirements by approximately 70\% relative to the original PyTorch FP32 versions.
For each evaluated architecture, statistical equivalence of Dice score between the PyTorch and quantised ONNX implementations was confirmed via TOST across the test set ($n = 300, \alpha = 0.05, \delta = 10^{-3}$), with all comparisons yielding $p < 0.001$. These results indicate that segmentation behaviour remains numerically stable despite the substantial reduction in precision and storage footprint.

\subsection{Performance across compressed models}
\begin{figure}[htb]
    \centering
    \includegraphics[width=0.95\linewidth]{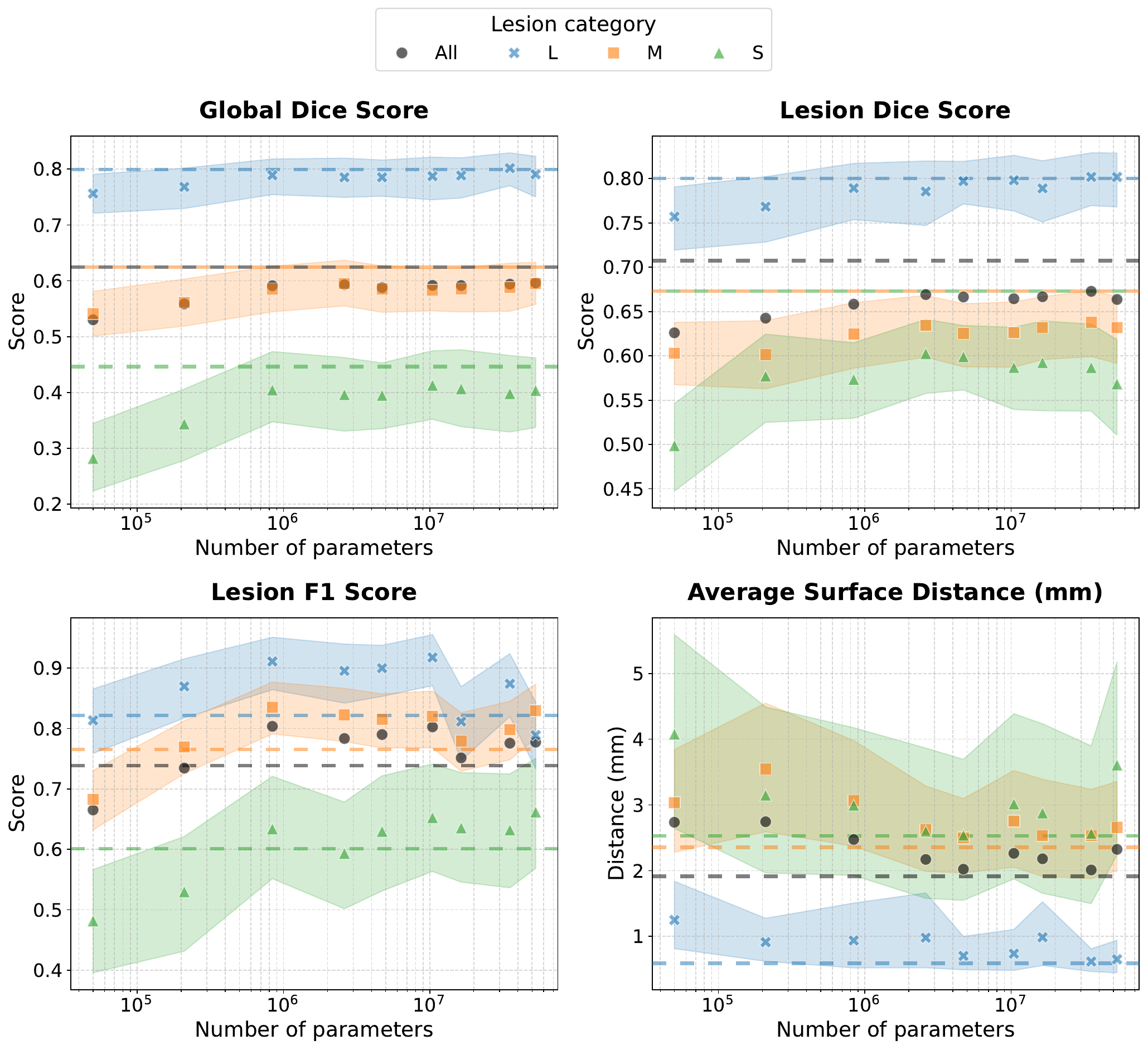}
    \caption{Segmentation (global Dice, lesion-wise Dice and ASD) and detection (lesion-wise F1) metrics across student networks. Results are shown globally (All) and stratified by lesion size (large, medium, small). Error bands represent 95\% confidence intervals. Dashed line indicates teacher model performance.}
    \label{fig:metrics_seg_det}
\end{figure}

The performance of the student networks across segmentation and detection metrics is illustrated in Fig.~\ref{fig:metrics_seg_det}. Segmentation metrics remained relatively stable across progressive model compression down to approximately 1M parameters (S7), with only moderate reductions in global and lesion-wise Dice scores compared with the teacher model. Below this threshold, performance deteriorated more rapidly. Performance remained dependent on lesion size, with small lesions consistently achieving lower scores and wider confidence intervals than medium and large lesions. Average surface distance progressively increased in the smallest student networks, reflecting reduced boundary delineation accuracy. We identify the S7 model as the most favourable trade-off, maintaining segmentation performance with a 2.1MB footprint, negligible compared with the 269MB application size.

\subsection{Runtime and energy efficiency}
\label{sec:powerExp}
Inference time and energy consumption per inference, corrected for baseline idle power, are summarised in Table~\ref{tab:performance_metrics}. Compared with the teacher model, the distilled S7 architecture substantially reduced both inference time and energy consumption across all hardware configurations. For FP32 inference, the strongest relative speed-up was observed on the CPU (15.2$\times$), followed by the iGPU (5.5$\times$) and dGPU (4.7$\times$). Energy reductions exceeded 90\% for all S7 configurations relative to the CPU FP32 teacher baseline. FP16 quantisation produced only minor changes in inference time compared with FP32 inference within the same architecture.

\begin{table}[H]
\caption{Performance and energy efficiency comparison across hardware targets for the teacher (\textbf{T}) and \textbf{S7} models. Energy reduction values are computed relative to the CPU FP32 teacher configuration.}
\label{tab:performance_metrics}
\centering
\small 
\begin{tabularx}{0.8\textwidth}{l *{8}{>{\centering\arraybackslash}X}}
\toprule
 & \multicolumn{4}{c}{\small\textbf{Inference Time (s)}} & \multicolumn{4}{c}{\small\textbf{Energy Reduction (\%)}} \\
 & \multicolumn{2}{c}{\small\textbf{FP32}} & \multicolumn{2}{c}{\small\textbf{FP16}} & \multicolumn{2}{c}{\small\textbf{FP32}} & \multicolumn{2}{c}{\small\textbf{FP16}} \\
\cmidrule(lr){2-5} \cmidrule(lr){6-9}
\textbf{} & \textbf{T} & \textbf{S7} & \textbf{T} & \textbf{S7} & \textbf{T} & \textbf{S7} & \textbf{T} & \textbf{S7} \\
\midrule
CPU   & 22.8 & 1.5 & 22.6 & 2.7 & --     & 95.8\% & 4.9\%  & 94.1\% \\
iGPU  & 32.2 & 5.9 & 37.0 & 4.7 & 55.5\% & 91.6\% & 54.9\% & 94.5\% \\
dGPU  & 5.6  & 1.2 & 4.8  & 1.3 & 86.2\% & 98.1\% & 92.6\% & 98.0\% \\
\bottomrule
\end{tabularx}
\end{table}

Packaging through the standalone C++/Qt deployment framework substantially reduced installer and application size compared with the original Python-based environment (Table~\ref{tab:packagesize}). The resulting Windows application required only 269~MB after installation, compared to roughly 2.8~GB in previous work~\cite{previouswork2025} and even more for legacy nnU-Net pipelines.

\begin{table}[H]
\caption{Installer and application size comparison. Model checkpoints excluded.}
\label{tab:packagesize}
\centering
\begin{tabular}{l>{\centering\arraybackslash}p{2cm}>{\centering\arraybackslash}p{2cm}>{\centering\arraybackslash}p{2cm}>{\centering\arraybackslash}p{2cm}}
\toprule
\textbf{} & \textbf{Baseline~\cite{previouswork2025} (MB)} & \textbf{Windows (MB)} & \textbf{macOS (MB)} & \textbf{Debian (MB)} \\
\midrule
Installer & 1220 & 136 & 201 & 142\\
App       & 2815 & 269 & 412 & 282\\
\bottomrule
\end{tabular}
\end{table}

\section{Discussion}
Our results show that 3D stroke lesion segmentation pipelines can be substantially reduced while preserving segmentation performance for clinical research workflows through knowledge distillation. Consistent with Di Matteo et al.~\cite{dimatteo2025stroke}, segmentation performance depended on lesion size: large lesions achieved higher Dice scores than medium and small ones. Lesion-wise detection (F1 score) was mostly unaffected by parameter reduction. Even in heavily compressed student networks, F1 remained stable. Overlap and delineation metrics remained relatively stable across progressive model compression down to approximately 1M parameters, with only moderate reductions compared with the teacher model. For smaller student architectures, segmentation quality deteriorated more rapidly. A secondary finding is that architectural distillation is the main source of runtime and energy gains, with \texttt{Float16} quantisation providing additional reductions in disk and memory requirements. We also tested NPU acceleration, but the SRAM buffer of the Intel\textsuperscript{\textregistered} AI Boost NPU could not accommodate large 3D fully convolutional networks. Only a heavily distilled model could be executed, without outperforming the iGPU, and initial compilation exceeded 30 min, making it impractical as the main inference backend. We therefore excluded NPU usage from the current deployment.

\section{Conclusion}
In this work, we extend StrokeSeg to StrokeSeg2, replacing the Python engine with a native C++ implementation and assessing model-level improvements. Using knowledge distillation, we compress a 102.3M-parameter teacher into a 0.84M-parameter student while maintaining segmentation performance. ONNX Runtime with \texttt{Float16} quantisation reduces model footprint while maintaining statistical equivalence with the PyTorch baseline and reducing model size by over 70\% relative to FP32. Hardware tests show architectural distillation to be the primary driver of runtime and energy gains, yielding over 90\% energy savings and an 84\% reduction in inference time across diverse devices. These improvements are packaged with a new software architecture into a self-contained, cross-platform C++/Qt application, eliminating Python and Docker dependencies and facilitating cross-device inference. Future work will integrate this solution into hospital research platforms, with deployment within Siemens Healthineers Open Recon interface as an initial step.

\begin{credits}
\paragraph{\textbf{Data and software availability.}}\label{sec:data_software_availability}
This study used retrospective MRI data from the public ATLAS v2.1 dataset~\cite{liew2018atlas} and pretrained models from~\cite{dimatteo2025stroke}. Per the dataset license, no additional ethical approval was required. The software is available at \url{https://strokeseg.readthedocs.io}, the code for reproduction of the experiments is available at \url{https://github.com/youwanM/StrokeSegLight}.

\paragraph{\textbf{\ackname}} This work was granted access to the HPC resources of IDRIS under the allocation 2026-AD011016064R1 made by GENCI. A total of 154 hours on H100 SXM hardware was required, with estimated emissions of 16 kgCO$_2$eq.

\paragraph{\textbf{\discintname}} The authors have no relevant financial or non-financial interests to disclose.

\end{credits}
\bibliographystyle{splncs04}
\bibliography{Paper-0002}

\end{document}